# Fluctuation-based Outlier Detection

Xusheng Du[1], Enguang Zuo[1], Zhenzhen He[1], Jiong Yu[1]

1. School of Information Science and Engineering, Xinjiang University, Urumqi 830046, China

**Abstract:** Outlier detection is an important topic in machine learning and has been used in a wide range of applications. Outliers are objects that are few in number and deviate from the majority of objects. As a result of these two properties, we show that outliers are susceptible to a mechanism called *fluctuation*. This article proposes a method called fluctuation-based outlier detection (FBOD) that achieves a low linear time complexity and detects outliers purely based on the concept of fluctuation without employing any distance, density or isolation measure. Fundamentally different from all existing methods. FBOD first converts the Euclidean structure datasets into graphs by using random links, then propagates the feature value according to the connection of the graph. Finally, by comparing the difference between the fluctuation of an object and its neighbors, FBOD determines the object with a larger difference as an outlier. The results of experiments comparing FBOD with seven state-of-the-art algorithms on eight real-world tabular datasets and three video datasets show that FBOD outperforms its competitors in the majority of cases and that FBOD has only 5% of the execution time of the fastest algorithm. The experiment codes are available at: https://github.com/FluctuationOD/Fluctuation-based-Outlier-Detection.

## 1. Introduction

The outlier detection problem can be defined as follows: given a dataset *X*, find objects that are considerably dissimilar, exceptional and inconsistent with respect to the remaining majority of objects [23]. Detecting outliers is important for many applications, such as financial fraud detection [19, 21], network analysis [2, 9], medical diagnosis [15, 11], intelligence agriculture [26, 28] and even the discovery of new stars in astronomy [20, 8].

The efficiency and effectiveness of most existing outlier detection methods, including distance-based and density-based methods, may be severely affected by increasing data volumes and dimensions due to the "curse of dimensionality". Clustering-based, classification-based, and autoencoder-based algorithms are usually byproducts of algorithms originally designed for purposes other than outlier detection. This leads to these methods often underperforming and detecting too few outliers. For isolation-based methods, when the number of objects is too large, normal objects interfere with the process of isolation and reduce the ability to detect outliers.

Outliers have two distinct properties: (i) they represent a very small proportion of the overall dataset; (ii) the feature value deviate significantly from the majority objects. If the neighbors of an outlier are overwhelmingly normal objects, the fluctuations, or degree of change in its feature values when the feature values of its neighbors are aggregated, will be more different from the fluctuations of its neighbors. Normal objects, which come from similar generative mechanisms, will have higher similarity between their feature values. When a normal object aggregates the feature values of its neighbors (the majority of which are normal objects), its fluctuations will be less different from those of its neighbors overall. Although fluctuation-based outlier detection is a very simple method, we show in this article that it is both effective and efficient in detecting outliers.

We summarize the main contributions of this work as follows:

1. Fluctuation-based outlier detection (FBOD) without employing any distance, density or isolation measure eliminates the major computational cost of distance calculation in all distance-based and density-based methods. Meanwhile, FBOD does not require any label or training process, which further improves the efficiency of detection.

2. FBOD has a linear time complexity with a small constant, which means FBOD has the capacity to scale up to handle large data sizes and extremely high-dimensional problems.

3. FBOD is able to perform outlier detection for both continuous valued data (e.g., common tabular data) and discrete valued data (images and videos).

## 2. Related work

The earliest research methods for outlier detection were statistical-based methods mainly designed to eliminate noise to improve the quality of data analysis. However, "one person's noise may be another person's signal" [22], and outliers are different from noise points, which are generally random errors in the distribution of the dataset and therefore do not have great value. In contrast, outliers have important significance to discovering the hidden mechanisms in a dataset [24].

Many algorithms have been designed to detect outliers. A taxonomy of various outlier detection methods divides them into six broad groups: statistical-based, distance-based, density-based, clustering-based, ensemble-based, and learning-based approaches. Note that it is possible that some methods can be categorized into more than one approach [17].

Statistical-based: In statistical-based outlier detection methods, the objects are sometimes modeled using a Gaussian distribution or regression methods, some objects can be labeled as outliers depending on the fitting degree with the distribution model. The straightforward statistical methods are the box plot, histogram and the Dixon-type test. Statistical-based approaches have a fast evaluation process once the models are built, but they are highly dependent on prior knowledge. Additionally, when faced with high-dimensional datasets, the processing time is sharply elevated [29].

Distance-based: Distance-based approaches are the most widely used and simple detection algorithms. Distance-based approaches identify outliers as points that are distant from their neighbors. Knorr and Ng define "a point $x$ in a dataset $X$ is a $DB(g, D)$ outlier if at least a fraction g of the points in $X$ lies at a greater distance than $D$ from $x$" [14]. Distance-based approaches have difficulty detecting outliers that lie within the normal object area and are computationally expensive in high-dimensional datasets [7].

Density-based: Density-based approaches consider objects of low-density regions as outliers. Well-known density-based approaches are the local outlier factor (LOF) [4], the connectivity-based outlier factor (COF) [12], and the local distance-based outlier factor (LDOF) [32]. Density-based approaches do not rely on assumed distributions to fit data, but in most cases, are more complicated and computationally expensive.

Clustering-based: Clustering-based outlier detection approaches usually take a two-step approach: grouping the data with clustering and analyzing the degree of deviation based on the clustering results. Their performance is highly dependent on the effectiveness of capturing the cluster structure of normal objects. In clustering-based methods, outliers are binary. There is no quantitative indication of the object's outlierness [30, 1, 16].

Ensemble-based: Ensemble-based approaches focus on combining the results of different types of outlier detection methods to generate more robust models that can detect outliers more effectively. They are more stable and give better predictive models, but for real-world datasets, outlier analysis can be very complex to evaluate, and selecting the right meta-detectors is a difficult task [34].

Learning-based: In most learning-based approaches, autoencoders play a central role, usually assuming that outliers are difficult to reconstruct by the autoencoder [10, 3, 6]. In recent years, some researchers have used GAN networks [18, 31], which can generate a large number of potential outliers, to balance the outlier detection problem into a binary classification problem. At the same time, many researchers have started to apply graph neural networks [13] to various types of domains due to their demonstrated power in processing graph data. In the domain of outlier detection, Wang proposed a framework for outlier detection on graphs incorporating GNN and OC-SVM [27]. Anshika proposed a GNN-based outlier detection model for social networks and e-mail [5]. Zhao optimized the loss function of GNN to better learn the feature representation of outliers in the graph [33]. The main advantage of learning-based methods is the ability to extract representative features from complex and high-dimensional data. The lack of accurate representation of normal and outlier boundaries also presents challenges for learning-based methods. Moreover, tuning the hyperparameters in learning-based approaches for optimal performances can be a challenging and time-consuming task.

## 3. Model

Let $X=\{x_1, x_2, x_3..., x_n\}$ be the given dataset of a $D$-variate distribution. The core idea of fluctuation-based outlier detection is as follows. The fluctuations produced by normal objects aggregating the feature values of their neighbors on themselves are similar to the fluctuations of their neighbors (the neighbors are basically normal objects). The fluctuations produced by outliers aggregating the feature values of their neighbors on themselves are completely different from the fluctuations of their neighbors. The fluctuation-based method compares the difference between the fluctuation of an object and its neighbors and determines the object with a larger difference as an outlier.

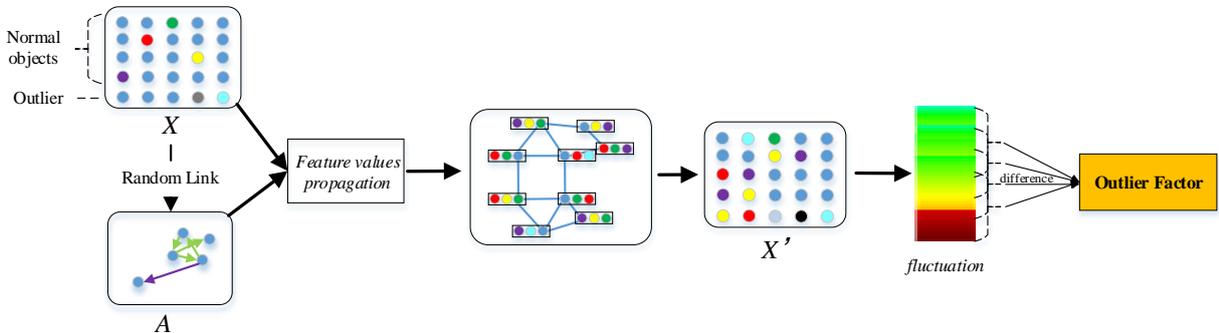

Fig 1. Illustration of the fluctuation-based outlier detection mechanism. FBOD compares the difference between the fluctuation of an object and its neighbors and determines the object with a larger difference as an outlier.

The fluctuation-based method consists of three parts: a) graph generation; b) feature value propagation; and c) outlier factor. We will describe these three parts in detail below.

## 3.1 Graph Generation

In the Euclidean structure dataset, objects are independent of each other, and there is no connection relationship. We propose a random link method to convert the original unconnected relationship into a connected relationship, called graph generation.

Definition 1: $k$ neighborhood of an object $x_i$

Randomly select $k$ objects from the dataset $X$; the set formed by these $k$ objects is named the $k$ neighborhood of an object $x_i$ and denoted as $N_k(x_i)$ ( $x_i \notin N_k(x_i)$ and $|N_k(x_i)|=k$ ). The process of constructing its $k$-neighborhood for any object is equivalent to subsampling the dataset $X$ once.

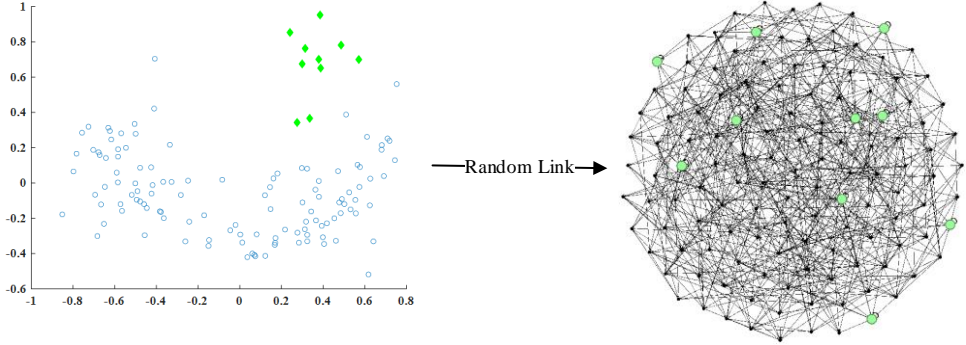

Fig 2. Illustration of the graph generation mechanism. The graph generated from real-world dataset wine ($k$=5); the solid green circles indicate the outliers. On the right side of Fig. 2, each object has 5 indegrees and is connected to itself.

Based on the fact that there are very few outliers in the dataset, even with the random link method, it is still possible to ensure that the $k$-neighborhoods of each object will contain mostly normal objects.

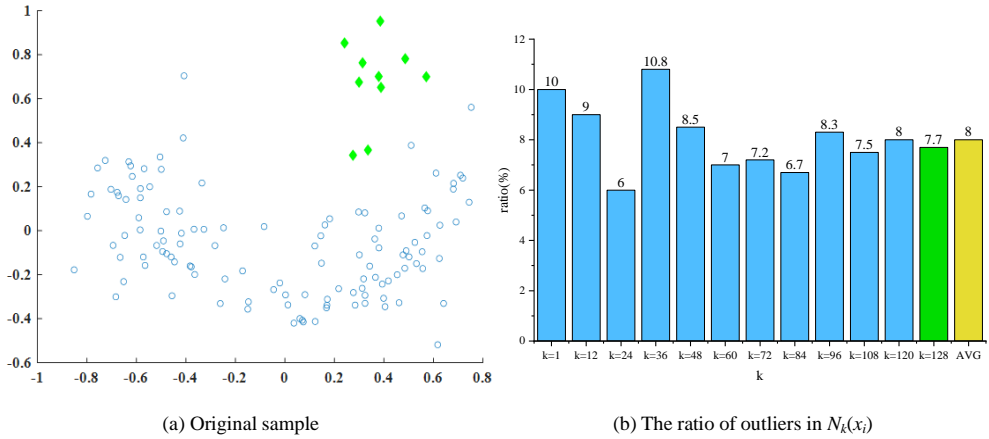

(a) Original sample  (b) The ratio of outliers in $N_k(x_i)$

Fig 3. Using the real-world dataset wine to describe the ratio of outliers in $N_k(x_i)$ is similar to the original ratio. Fig 3(a) includes 129 objects with 10 outliers, and the rate of outliers is 7.75%. Figure 3(b) shows that by randomly selecting $k$ objects from Fig 3(a) and constructing the neighborhood of $x_i$, the average rate of outliers in $N_k(x_i)$ is 8% (when $k$ takes different values, the average value is taken of 10 executions).

Graph generation consists of three main steps: 1) construct $N_k(x_i)$; 2) construct a directed edge from the object's neighbors to itself; 3) store the constructed graph in the adjacency matrix $A$.

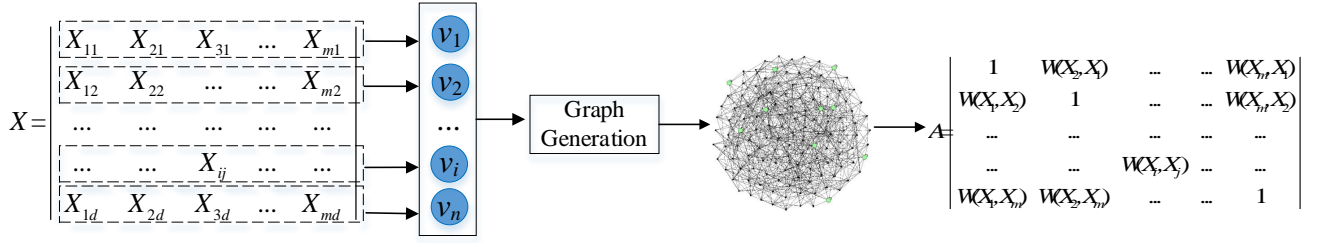

Fig 4. Graph generation

| **Algorithm1** Graph Generation |
|---|
| **Input:** Given dataset *X*, the number of neighbors *k*. |
| **Output**: Generated graph (adjacency matrix *A*) |
| 1. [*m*, *n*]=size(*X*); |
| 2. for i=1:*m* |
| 3.     $N_k(i,:)$ = randperm (*m*, *k*); |
| 4. end |
| 5. Construct a directed edge from the object's neighbors to the object itself. |
| 6. Set the weights on the directed edges and diagonal value of *A* to 1. |
| 7. **return** *A*. |

### 3.2 Feature Value Propagation and Fluctuation

Let the graph generation from *X* be *A*. Design the feature value propagation as:

$$X' = X * A \quad (1)$$

We provide an example to illustrate feature value propagation in more detail. Let $X = \begin{bmatrix} x_{11} & x_{21} \\ x_{12} & x_{22} \end{bmatrix}$. Each column in *X* represents an object, and each object has two features.

$A = \begin{bmatrix} 1 & W(x_2, x_1) \\ W(x_1, x_2) & 1 \end{bmatrix}$ , $X * A = \begin{bmatrix} x_{11}*1 + x_{21}*W(x_1, x_2) & x_{11}*W(x_2, x_1) + x_{21}*1 \\ x_{12}*1 + x_{22}*W(x_1, x_2) & x_{12}*W(x_2, x_1) + x_{22}*1 \end{bmatrix}$ . After

feature value propagation, object *x* contains not only its own feature value but also the feature value passed by its neighbors.

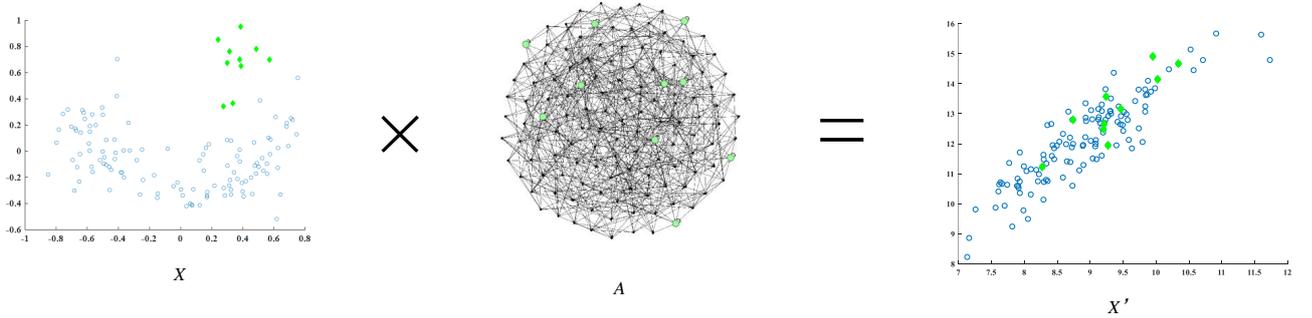

Fig 5. An illustrate graph of feature values propagation. The rightmost side of Fig. 4 shows the distribution of *X* after feature value propagation. When the object aggregates the feature values of other objects, its own feature values change, resulting in a significant change in the overall data distribution.

Definition 2. Fluctuation

Let $x_i'$ denote $x_i$ after feature value propagation; then:

$$fluctuation(x_i) = \sum_{d=1}^{D} \frac{x_{id}}{x_{id}'} = \sum_{d=1}^{D} \frac{x_{id}}{x_{id} + \sum_{k=1}^{k} N_k(x_i)_d} \quad (2)$$

In Equation (2), $N_k(x_i)_d$ denotes the feature value of the $k$th neighbor of object $x_i$ in the $d$th dimension. From Section 3.1, it is clear that the majority of those contained in $N_k(x_i)$ are normal objects. Since the feature values of normal objects are similar and the feature values of outlier and normal objects have large differences, we can deduce that:

$$|\sum_{k=1}^{k} N_k(x_i)_d - k * x_{id}| \approx 0 \quad if \quad x_i \in normal\ object$$
$$|\sum_{k=1}^{k} N_k(x_i)_d - k * x_{id}| >> 0 \quad if \quad x_i \in outlier \quad (3)$$

Combining Equation (2) with Equation (3):

$$fluctuation(x_i) \approx \sum_{d=1}^{D} \frac{x_{id}}{x_{id} + k \times x_{id}} = \frac{1}{1+k} \quad if \quad x_i \in normal\ object$$

$$fluctuation(x_i) = \sum_{d=1}^{D} \frac{x_{id}}{x_{id} + \sum_{k=1}^{K} N_k(x_i)} << \frac{1}{1+k} \quad if \quad x_i \in outlier\ and\ \sum_{k=1}^{K} N_k(x_i)_d >> k \times x_{id} \quad (4)$$

$$fluctuation(x_i) = \sum_{d=1}^{D} \frac{x_{id}}{x_{id} + \sum_{k=1}^{K} N_k(x_i)} >> \frac{1}{1+k} \quad if \quad x_i \in outlier\ and\ \sum_{k=1}^{K} N_k(x_i)_d << k \times x_{id}$$

If $x_i$ belongs to a normal object, its fluctuation value is approximately equal to $1/(1+k)$; if $x_i$ belongs to an outlier, the fluctuation value will be significantly different from that of the normal objects.

We use a two-dimensional synthetic dataset to demonstrate significant differences in the fluctuations of outlier and normal objects:

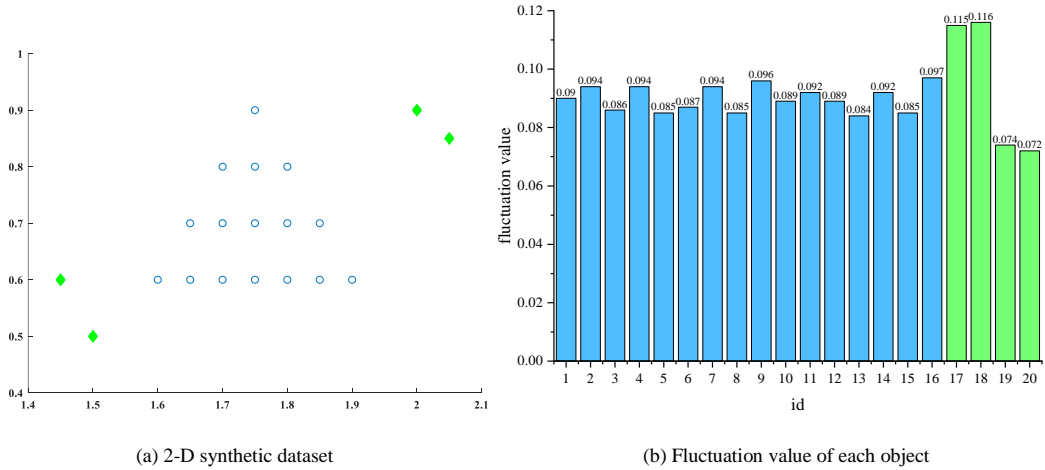

(a) 2-D synthetic dataset    (b) Fluctuation value of each object

Fig 6. 2D synthetic dataset and fluctuation value of each object. Sixteen normal objects are represented by blue hollow circles, and four outlier points are represented by solid green diamonds.

After the objects in the 2D synthetic dataset are propagated by feature values, their fluctuation values are significantly different due to two different generation mechanisms of outliers and normal objects. The fluctuation values of the two outliers located in the upper right corner are significantly higher than those of the normal objects, while the fluctuation values of the outliers located in the lower left corner are significantly lower than those of the normal objects. Based on the difference in fluctuation values, the outlier factor of the objects can be further calculated in the downstream task.

### 3.3 Outlier Factor

The outlier factor of $x_i$ is defined as:

$$OF(x_i)=\sum_{t=1}^{T} \sum_{x_j \in N_k(x_i)} | fluctuation(x_i) - fluctuation(x_j)| \quad (5)$$

In Equation (5), $T$ denotes the number of generated graphs. Since a random connection is used in the graph generation process, multiple graphs are used to ensure the stability of the FBOD. The outlier factor of object $x_i$ captures the degree to which we call $x_i$ an outlier. The larger the OF value is, the more likely it is that the object is an outlier; the smaller the OF value is, the more likely it is that the object is normal.

**Algorithm2** Outlier Factor

**Input:** Given dataset $X$, Adjacency matrix $A$, the number of neighbors $k$, the number of outliers $p$

**Output**: The set of outliers $O$

1. Use Equation (3) to calculate the $OF$ of the object.
2. [value, index] = $Sort$($OF$, 'descend');
3. $O$ = index (1:$p$,:);
4. return $O$.

### 3.4 Time Complexity Analysis

The steps of FBOD to detect outliers contain a total of 3 parts: graph generation, feature value propagation and fluctuations, and calculation of outlier factors. We analyze the time complexity of each part in detail:

a). Graph generation: The time complexity of randomly selecting $k$ neighbors for $n$ objects in the dataset is $O(k * n)$.

b). Feature value propagation: The time complexity of each object in $X$ aggregating the feature values of its neighbors is $O(k * n)$.

c). Fluctuation: The original feature values of each object are compared with the feature values after feature value propagation with a time complexity of $O(n)$.

d). Outlier factor: The time complexity of comparing the fluctuating values of each object with the feature values of its neighbors is $O(k * n)$.

Since $k$ is a constant, the overall time complexity of the FBOD algorithm is $O(n)$.

## 4. Experiments

To verify the effectiveness of the FBOD method, we compare the FBOD with several state-of-art algorithms in real datasets. The source code of our model is implemented in MATLAB 2019A. The hardware environment for the experiments is an Intel(R) Core(TM) i7-7700 3.60 GHz CPU with 16 GB of RAM. The operating system environment is Microsoft Windows 10 Professional.

### 4.1 The Summary of Datasets and Compared Algorithms

ODDS (http://odds.cs.stonybrook.edu/) openly provides access to the collection of outlier

detection datasets with ground truths in various real-world fields. We use the multidimensional point datasets in the ODDS and remove duplicate objects.

Table 1. Dataset statistics

| Dataset | # of records | # of features | # of outliers | Outlier ratio |
|---|---|---|---|---|
| breastw | 683 | 9 | 239 | 34.9% |
| wbc | 377 | 30 | 20 | 5.3% |
| wine | 129 | 13 | 10 | 1.2% |
| heart | 267 | 44 | 55 | 20.6% |
| vowels | 1452 | 12 | 46 | 3.1% |
| lympho | 148 | 18 | 6 | 4.1% |
| pima | 768 | 8 | 268 | 34.8% |
| glass | 213 | 9 | 9 | 4.2% |

We selected five different types of seven state-of-the-art outlier detection algorithms for comparison experiments with the proposed FBOD. These algorithms are common types in the outlier detection field, and are used as comparison algorithms in most related literature. To compare the performance of each algorithm fairly, all algorithms are implemented in MATLAB 2019A.

Table 2. Comparison algorithm statistics

| Type of algorithm | Acronym of algorithm |
|---|---|
| Neuron network-based | AE |
| Graph-based | CutPC |
| Local outlier factor-based | LOF, COF |
| Clustering-based | K-means, OPTICS |
| Isolation-based | IForest |

Due to the large variety and number of algorithms in the comparison experiments, the parameter settings for each type of algorithm are different. Therefore, Table 3 is used to describe in detail the parameter settings of each algorithm in the experiments.

Table 3. Parameter Setting

| Algorithms | $k$(Number of nearest neighbors) | Number of Graphs | Learning rate | Number of iterations | Number of layers | $xi$(relative decrease in density) | minpts(Number of points required to form a cluster) | Number of isolation trees & subsample size |
|---|---|---|---|---|---|---|---|---|
| FBOD | 2~100 | 1~5 | \ | \ | \ | \ | \ | \ |
| AE | \ | | 0.0001~0.002 | 10~100 | 3 | \ | \ | \ |
| CutPC | \ | \ | \ | \ | \ | \ | \ | \ |
| LOF | 2~100 | \ | \ | \ | \ | \ | \ | \ |
| COF | 2~100 | \ | \ | \ | \ | \ | \ | \ |
| K-means | 2~100 | \ | \ | \ | \ | \ | \ | \ |
| OPTICS | \ | \ | \ | \ | \ | 0.1 | 2~100 | \ |
| IForest | \ | \ | \ | \ | \ | \ | \ | 100, 56~256 |

## 4.2 Evaluation Techniques

In real-world applications, ground truth outliers are generally rare. The receiver operating characteristic (ROC) curve, which captures the trade-off between sensitivity and specificity, has been adopted by a large proportion of studies in this area. The area under the curve (AUC), which ranges from 0-1, characterizes the area under the ROC curve. An algorithm with a large AUC value is preferred [25]. We choose the AUC, accuracy (ACC), detection rate (DR), and false alarm rate (FAR) as the algorithm performance evaluation metrics. Higher AUC, ACC, and DR values and lower FAR and execution time indicate better performance.

Let TP be the number of outliers correctly marked as outliers by the algorithm (TOP-$p$); TN be the number of normal objects correctly marked as normal by the algorithm; FP be the number of normal objects incorrectly marked as outliers by the algorithm; FN be the number of outliers that the algorithm incorrectly marks as normal objects. The calculation method of each evaluation techniques are shown in Algorithm 3.

**Algorithm 3** Evaluation Techniques

1. Let $n_o$ be the number of true outliers.
2. Let $n_n$ be the number of true normal objects.
3. Rank all objects according to their outlier factors in descending order.
4. Let $S$ be the sum of ranking of the actual outliers, $S = \sum_{i=1}^{n_o} r_i$, where $r_i$ is the rank of the $i$th outlier in the ranked list.
5. $AUC = \dfrac{S - (n_o^2 + n_o)/2}{n_o n_n}$
6. $ACC = \dfrac{TP + TN}{TP + TN + FP + FN}$
7. $DR = \dfrac{TP}{TP + FN}$
8. $FAR = \dfrac{FP}{TN + FP}$

## 4.3 Experimental on Real-World Tabular Datasets

We judged the $p$ objects (the size of $p$ is equal to the number of true outliers in the dataset) with the highest outlier factor determined by the FBOD as outliers and compared them with the labels. Fig 7 shows the fluctuation values of each object as determined by FBOD. The left side of Fig 7 shows the original data distribution after dimensionality reduction using principal component analysis (PCA), and the right side shows the fluctuation values of each object determined by FBOD.

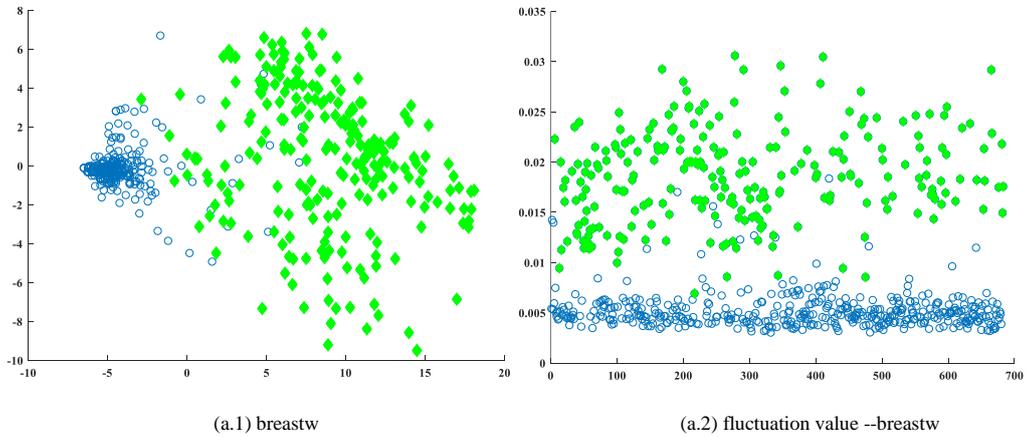

(a.1) breastw            (a.2) fluctuation value --breastw

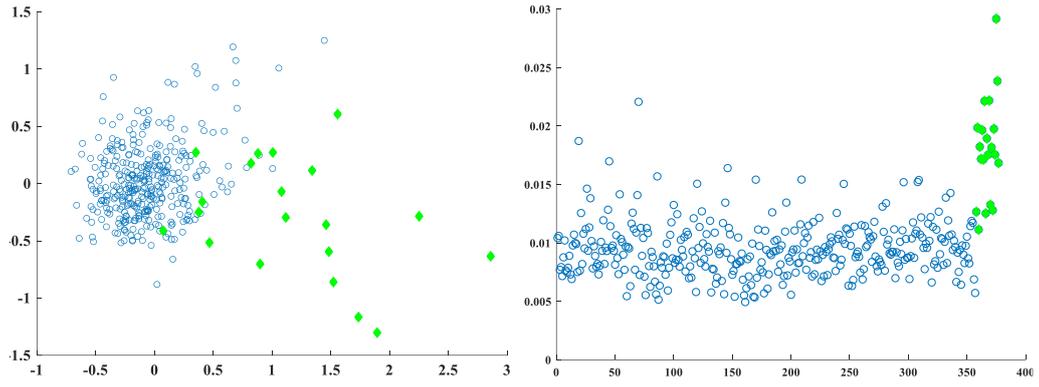

(b.1) wbc           (b.2) fluctuation value --wbc

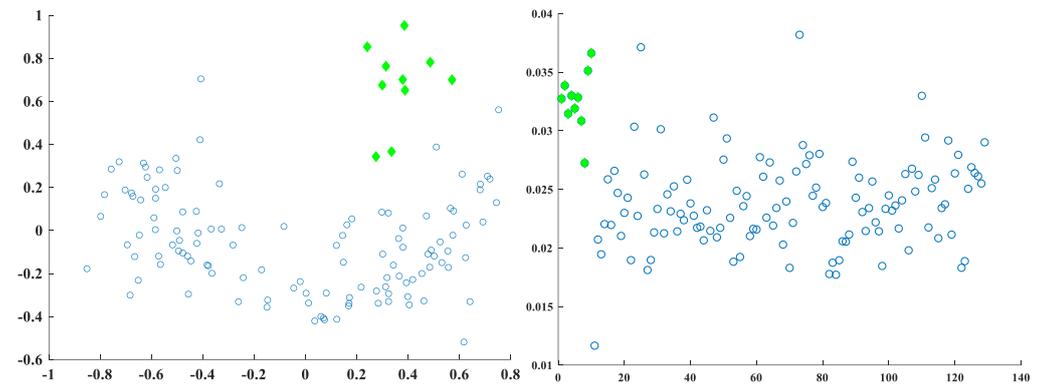

(c.1) wine           (c.2) fluctuation value --wine

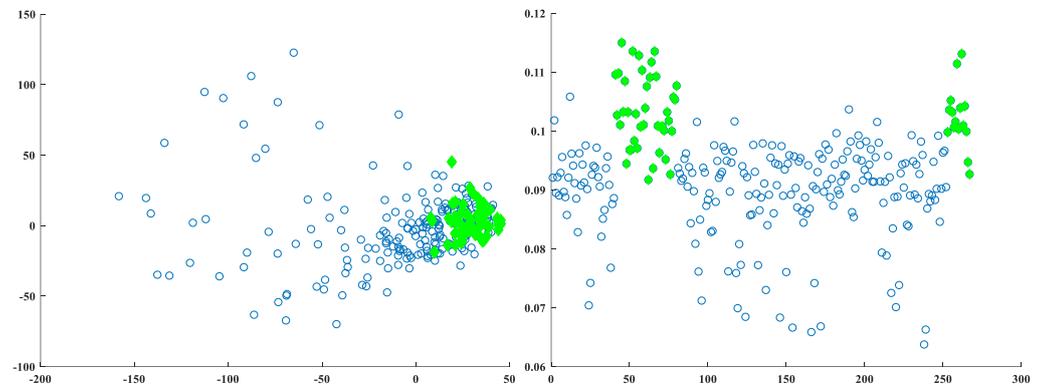

(d.1) heart           (d.2) fluctuation value --heart

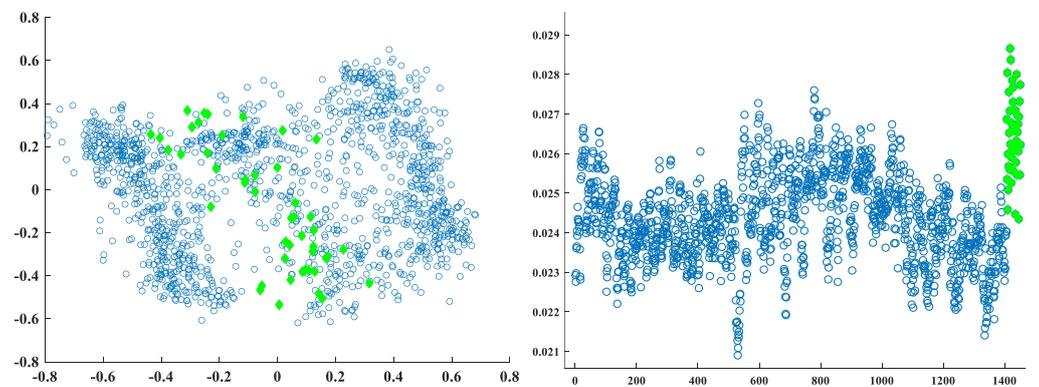

(e.1) vowels           (e.2) fluctuation value --vowels

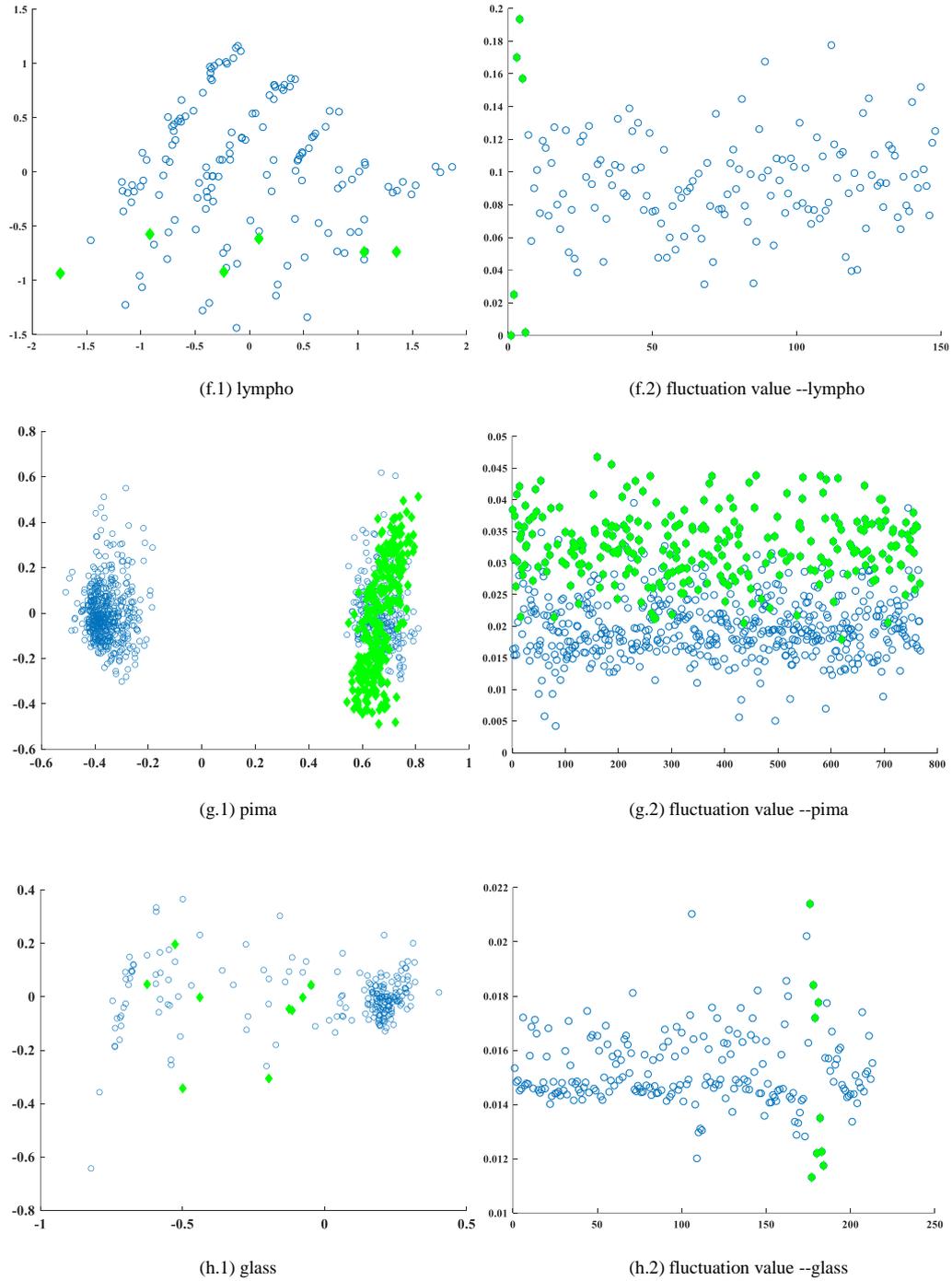

Fig 7. The original data distribution after PCA dimensionality reduction. The blue hollow circles represent normal objects, the green solid diamonds represent the true outliers in each dataset (left) and fluctuation values of each object in datasets, and the green solid diamonds represent the fluctuation values of true outliers in each dataset (right).

Fig 7 shows the breastw, wbc, wine, heart, vowels, lympho, pima, and glass datasets from top to bottom. There are many outliers in the original dataset that are mixed in the normal object area on the left. These outliers are usually difficult to detect by distance-based, density-based and clustering-based algorithms. On the right side of Fig 7, after the feature value propagation, the fluctuation values of normal objects and outliers are significantly different; therefore, true outliers can be detected in downstream tasks based on the difference in fluctuation values between objects.

For each algorithm on each dataset, we adjusted the parameters and ran 20 experiments. Thus,

a total of 1280 experiments were conducted, and the best results of each algorithm on each dataset were selected as the final evaluation of the algorithm performance. The experimental results of our proposed method and seven comparison algorithms on eight datasets are shown in Table 4.

Table 4. Experimental results on real-world datasets. Values marked in bold are ranked by the top 2 in the dataset.

| Datasets | Time(seconds) | | | | | | | |
|---|---|---|---|---|---|---|---|---|
| | FBOD | AE | CutPC | LOF | COF | K-means | OPTICS | IForest |
| breastw | **0.0026** | 0.2353 | 0.1232 | 0.1539 | 1.2333 | 0.1240 | **0.1215** | 0.1692 |
| wbc | **0.0028** | 0.1404 | **0.0370** | 0.2016 | 0.5517 | 0.0824 | 0.0464 | 0.1372 |
| wine | **0.0004** | 0.0421 | 0.0325 | 0.0573 | 0.1698 | **0.0062** | 0.0096 | 0.1421 |
| heart | **0.0023** | 0.0990 | 0.0375 | 0.0798 | 0.3482 | **0.0200** | 0.0250 | 0.2282 |
| vowels | **0.0107** | 0.3380 | 0.1847 | 0.4356 | 25.981 | **0.1272** | 0.1757 | 0.1859 |
| lympho | **0.0004** | 0.0473 | 0.3153 | 0.1057 | 0.5457 | 0.2353 | **0.0113** | 0.1566 |
| pima | **0.0030** | 0.1944 | 0.0776 | 0.2131 | 23.531 | **0.0360** | 0.0594 | 0.1722 |
| glass | **0.0006** | 0.0626 | 0.0162 | 0.0820 | 2.3947 | **0.0076** | 0.0138 | 0.1285 |

(a) Actual execution time

| Datasets | AUC | | | | | | | |
|---|---|---|---|---|---|---|---|---|
| | FBOD | AE | CutPC | LOF | COF | K-means | OPTICS | IForest |
| breastw | **0.9903** | 0.8900 | 0.9238 | 0.8131 | 0.6962 | 0.8814 | 0.8522 | **0.9400** |
| wbc | **0.9794** | 0.9092 | 0.9529 | 0.9555 | 0.9328 | 0.9372 | 0.9405 | **0.9761** |
| wine | **0.9672** | **0.9555** | 0.5286 | 0.9244 | 0.779 | 0.7790 | 0.8542 | 0.9101 |
| heart | **0.5133** | **0.4901** | 0.1910 | 0.2420 | 0.2314 | 0.3652 | 0.3321 | 0.2651 |
| vowels | **0.9216** | 0.8007 | **0.9889** | 0.8480 | 0.8180 | 0.5554 | 0.9078 | 0.6894 |
| lympho | **0.9061** | 0.5434 | **0.9812** | 0.7899 | 0.7840 | 0.8847 | 0.8635 | 0.8865 |
| pima | 0.7817 | **0.9467** | 0.7798 | 0.4900 | 0.7363 | **0.9510** | 0.8893 | 0.6700 |
| glass | 0.7664 | 0.5697 | **0.8399** | 0.7941 | **0.8769** | 0.4733 | 0.7216 | 0.7118 |

(b) AUC performance

| Datasets | ACC(%) | | | | | | | |
|---|---|---|---|---|---|---|---|---|
| | FBOD | AE | CutPC | LOF | COF | K-means | OPTICS | IForest |
| breastw | **96.77** | 94.43 | 94.43 | 68.66 | 91.50 | 68.66 | 68.66 | **96.48** |
| wbc | **97.34** | 94.69 | 95.22 | **95.75** | 94.69 | 94.69 | 95.22 | **95.75** |
| wine | **93.79** | **92.24** | 84.49 | 90.69 | 84.49 | 87.59 | 89.14 | **92.24** |
| heart | **67.04** | 61.04 | 59.55 | 61.04 | 61.04 | **65.54** | 64.04 | 59.55 |
| vowels | **96.28** | 94.35 | **98.07** | 95.73 | 95.86 | 94.07 | 95.86 | 94.76 |
| lympho | **95.94** | 93.24 | **97.29** | 94.59 | 94.59 | 94.59 | 94.59 | 94.59 |
| pima | 73.69 | **85.15** | 70.05 | 70.05 | 62.50 | **95.83** | 82.29 | 72.91 |
| glass | **94.36** | 92.48 | 92.48 | 92.48 | **93.45** | 92.48 | 92.48 | 92.48 |

(c) Accuracy performance (%)

| Datasets | DR(%) | | | | | | | |
|---|---|---|---|---|---|---|---|---|
| | FBOD | AE | CutPC | LOF | COF | K-means | OPTICS | IForest |
| breastw | **95.39** | 92.05 | 92.05 | 55.23 | 87.86 | 55.23 | 55.23 | **94.97** |
| wbc | **75** | 50 | 55.00 | **60** | 50 | 50 | 55.00 | **60** |
| wine | **70** | **50** | 0 | 40 | 0 | 20.0 | 30 | **50** |
| heart | **20** | 5.45 | 1.81 | 5.45 | 5.45 | **16.36** | 12.72 | 1.81 |
| vowels | **41.30** | 10.86 | **69.56** | 32.60 | 34.78 | 6.52 | 34.78 | 17.39 |

| Datasets | DR(%) | | | | | | | |
|---|---|---|---|---|---|---|---|---|
| | FBOD | AE | CutPC | LOF | COF | K-means | OPTICS | IForest |
| lympho | **50** | 16.67 | **66.67** | 33.33 | 33.33 | 33.33 | 33.33 | 33.33 |
| pima | 62.31 | **78.73** | 57.08 | 57.08 | 45.79 | **94.02** | 74.62 | 61.19 |
| glass | **33.33** | 11.11 | 11.11 | 11.11 | **22.22** | 11.1 | 11.11 | 11.11 |

(d) Detection rate performance (%)

| Datasets | FAR(%) | | | | | | | |
|---|---|---|---|---|---|---|---|---|
| | FBOD | AE | CutPC | LOF | COF | K-means | OPTICS | IForest |
| breastw | **2.47** | 4.27 | 4.27 | 24.09 | 6.53 | 24.09 | 24.09 | **2.70** |
| wbc | **1.40** | 2.80 | 2.52 | **2.24** | 2.80 | 2.80 | 2.52 | **2.24** |
| wine | **2.52** | **4.20** | 8.40 | 5.04 | 8.40 | 6.72 | 5.88 | **4.20** |
| heart | **20.75** | 24.52 | 25.47 | 24.52 | 24.52 | **21.69** | 22.64 | 25.47 |
| vowels | **1.92** | 2.91 | **0.99** | 2.20 | 2.13 | 3.05 | 2.13 | 2.70 |
| lympho | **2.11** | 3.52 | **1.40** | 2.81 | 2.81 | 2.81 | 2.81 | 2.81 |
| pima | 20.20 | **11.40** | 23.00 | 23.00 | 30.00 | **3.20** | 13.6 | 20.80 |
| glass | **2.94** | 3.92 | 3.92 | 3.92 | **3.41** | 3.92 | 3.92 | 3.92 |

(e) False alarm rate (%)

In the eight real-world datasets, we compared the execution times of the eight algorithms. Observing Table 4, we can conclude that the execution time of the FBOD is much lower than that of the comparison algorithms. The average execution time of FBOD is only 5% of the average execution time compared to the OPTICS algorithm, which has the fastest average execution time. Furthermore, FBOD accounts for only four ten-thousandths of its execution time compared to the COF algorithm, which has the slowest average execution time.

The main reasons for the extremely low time overhead of FBOD are that (1) it does not require the calculation of distances or densities, and (2) it only calculates the difference between the fluctuations of the object itself and the fluctuations of its neighbors and does not involve the calculation of relationships with all objects in the dataset.

From Table 4(b, c, d, e), several observations can be obtained:

(i) FBOD obtains the best results in four datasets: breastw, wbc, wine, and heart. Since the heart dataset has 44 dimensions, the curse of dimensionality caused a near failure of the distance- or density-based methods.

(ii) FBOD achieved the next best detection results in the vowels and lympho datasets, and its performance was still significantly higher than most of the comparison algorithms.

(iii) The average AUC value of FBOD on eight real datasets is 0.85, which is 6% higher than the next highest OPTICS method. The average AUC value of FBOD is 10% higher than that of the isolated forest algorithm. The effectiveness of FBOD is demonstrated by the excellent detection results achieved in different types of outlier detection tasks.

### 4.4 Experiments on Video Data

The UCSD outlier detection dataset (http://www.svcl.ucsd.edu/projects/anomaly/dataset.htm) was acquired with a stationary camera mounted at an elevation, overlooking pedestrian walkways. In the normal setting, the video contains only pedestrians. Outliers are due to either:

a): The circulation of nonpedestrian entities in walkways.

b): Anomalous pedestrian motion patterns.

Commonly occurring anomalies include bikers, skaters and small carts. All outliers are naturally occurring, i.e., they were not staged for the purposes of assembling the dataset. We constructed 3 datasets based on the UCSD library, and each dataset contained 60 normal images and 3 outliers. Finally, each dataset is transformed into a 37604*63 matrix *X*. Each column of *X* represents an object (a picture), and each row corresponds to the value of a pixel point at a certain location.

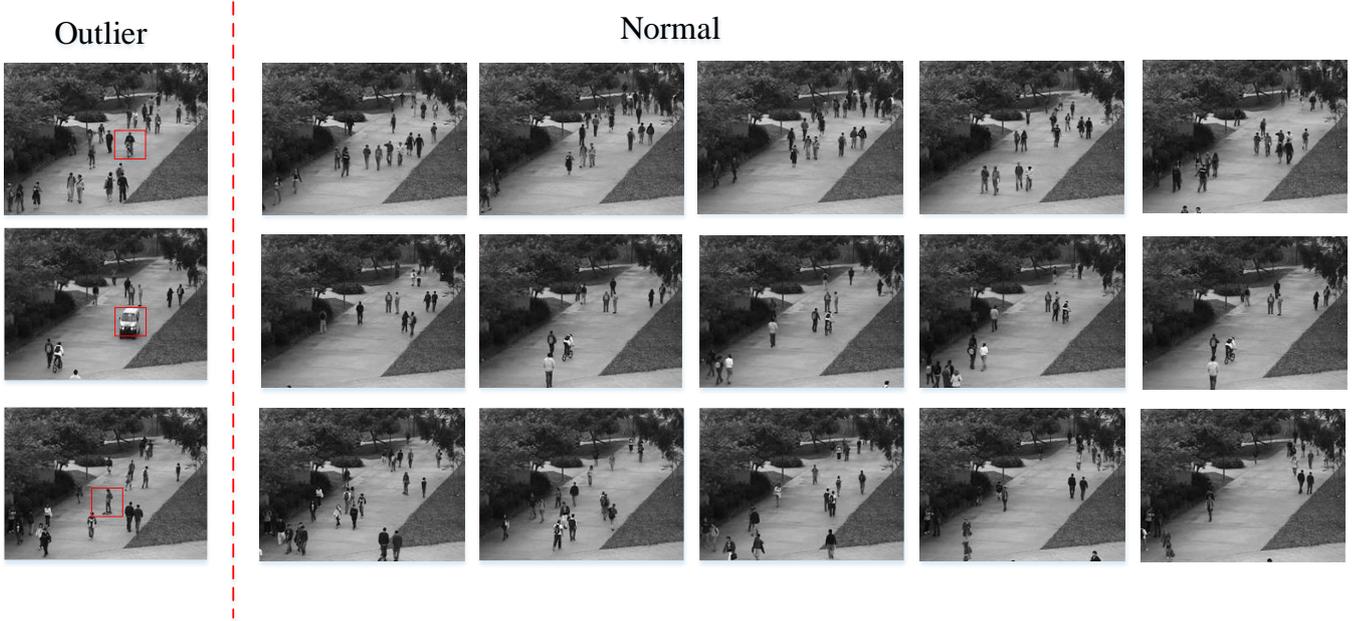

Fig 8. Video outlier detection dataset. The outliers on the leftmost are anomalous events on the sidewalk with bikers, small carts and skaters, and we have highlighted the cause of the anomaly with a red rectangle.

All the outlier detection algorithms in Section 4.1 cannot work due to the high dimensionality of video data. The FBOD algorithm has features that make it possible to achieve effective detection in arbitrarily high dimensions. FBOD selects those images with the largest differences in pixel values as outliers based on the differences in pixel values between images.

Table 5. Detection results in video datasets

| Datasets | AUC | Accuracy (%) | Detection Rate (%) | False Alarm Rate (%) | Execution time(second) |
|---|---|---|---|---|---|
| bike | 1 | 100 | 100 | 0 | 0.0149 |
| car | 1 | 100 | 100 | 0 | 0.0122 |
| skateboard | 0.98 | 96.8254 | 66.6667 | 1.6667 | 0.0132 |

FBOD successfully detects all outliers in the biker and small cart dataset on the sidewalk. In the skateboard dataset, only two of the three outliers are detected because the skaters are less different from the pedestrians. In terms of execution time, the FBOD takes only 0.013 seconds to complete the detection process and has the potential to handle real-world video anomaly detection tasks.

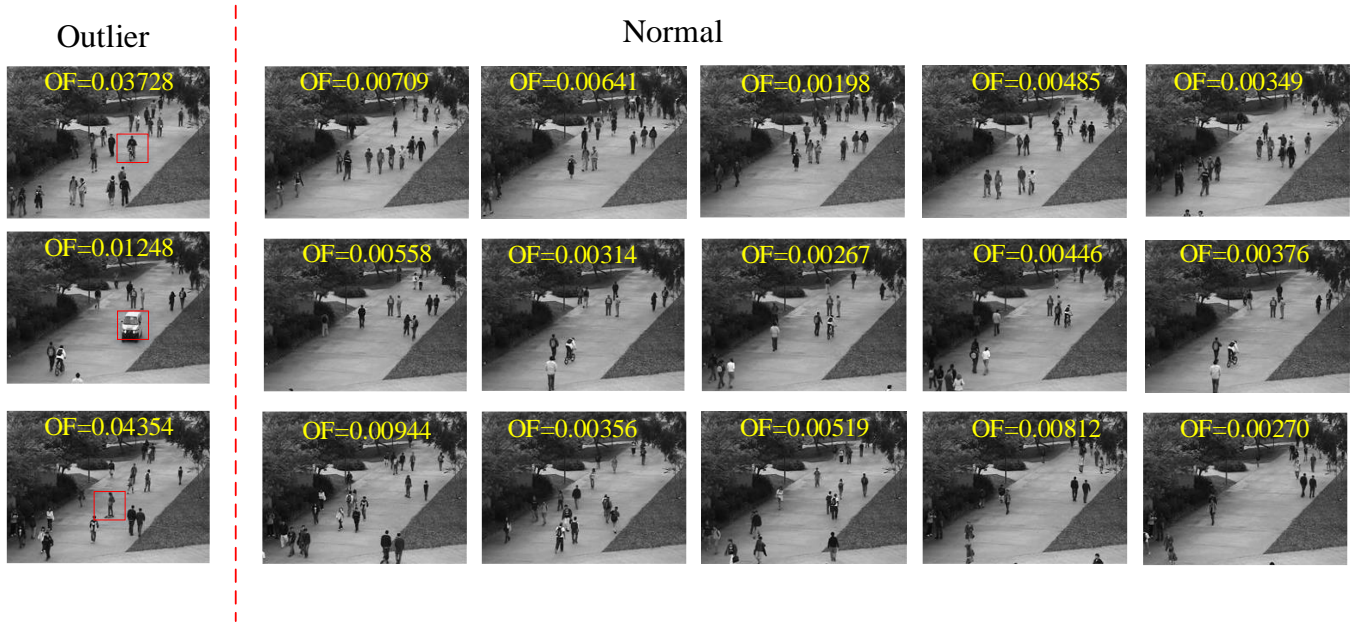

Fig. 9. Visualization of FBOD detection results. The OF value of the leftmost outlier is significantly higher than that of the normal object.

## 4.5 Research on the Influence of Parameters on the Detection Performance of FBOD

We conducted 20 experiments on each real-world dataset to investigate the effects of the number of neighbors k and the graph number $T$ on the performance of the FBOD. The experimental results are shown below.

### 4.5.1 FBOD performance against various neighbor's $k$

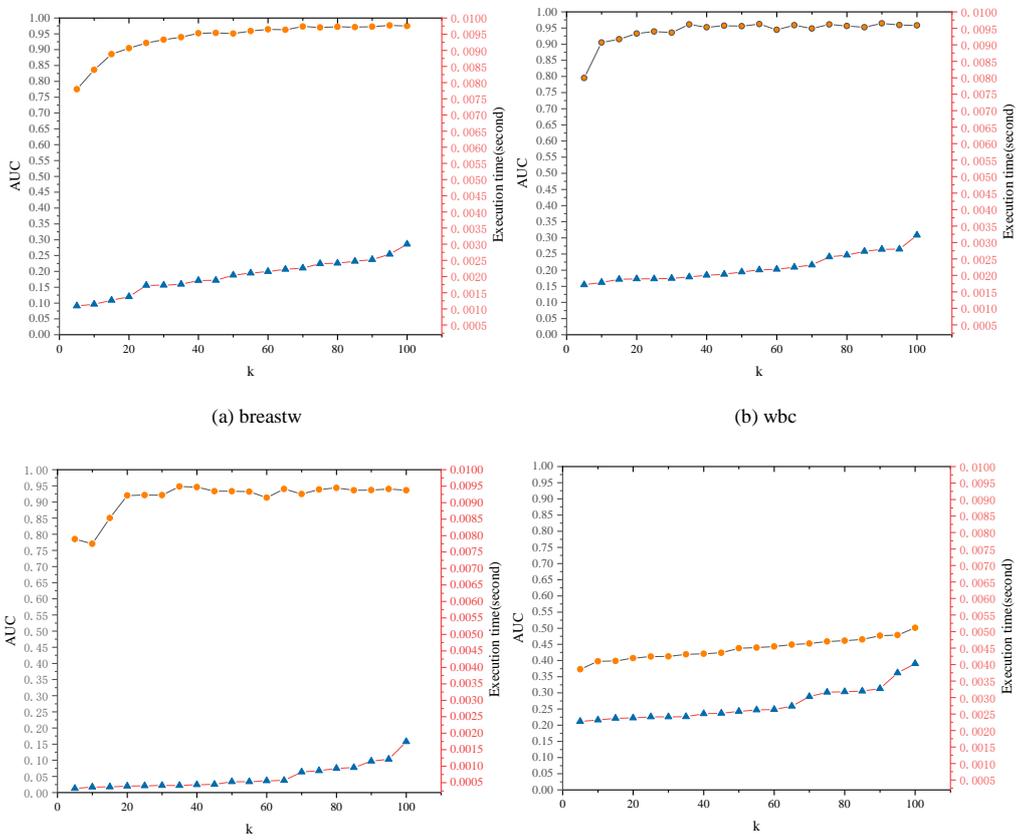

(a) breastw  (b) wbc

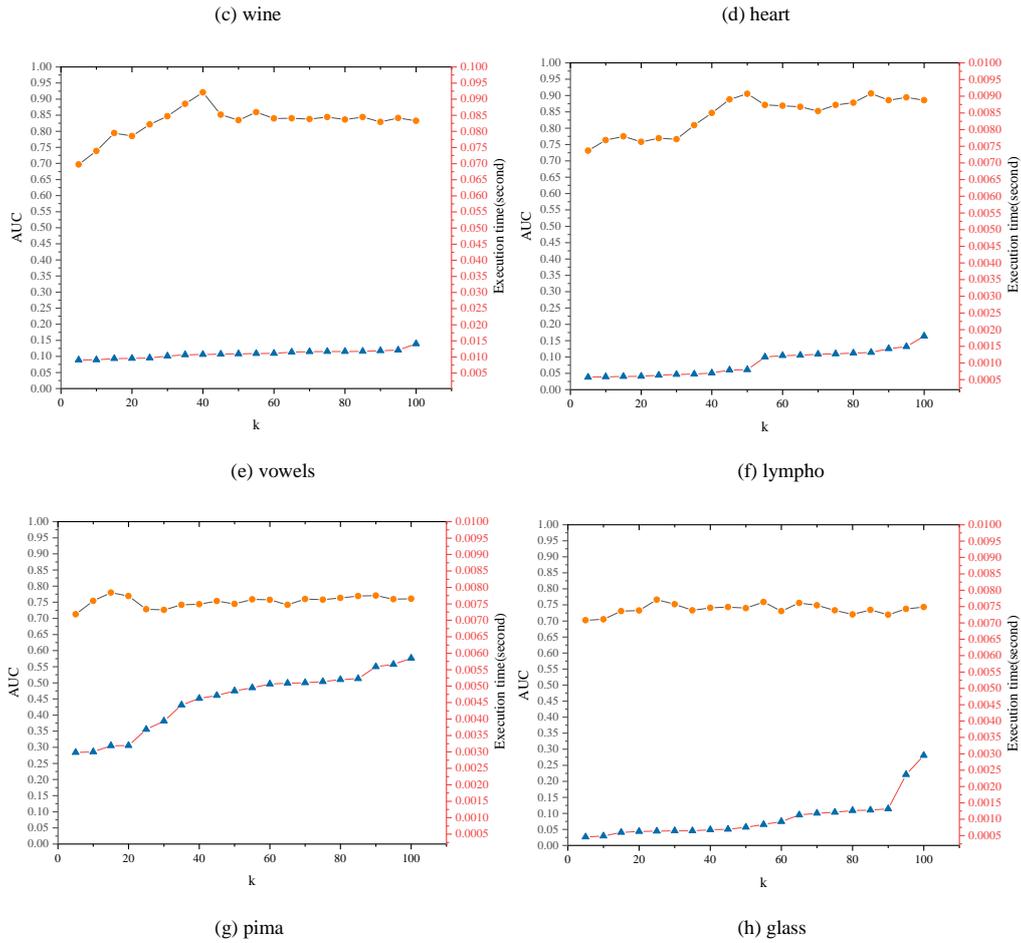

Fig 10. FBOD AUC performance (y1-axis) and execution time (y2-axis) versus the different neighbors' $k$ (x-axis).

Parameter $k$ ranges from 5 to 100, and the value of $k$ increases by 5 each time (graph number $T$=2). Fig. 7 shows that when $k$ gradually increases, the AUC of FBOD also gradually increases. After reaching the highest AUC, $k$ continues to increase, and the AUC value tends to stabilize. At the same time, when $k$ gradually increases, the execution time of FBOD gradually increases.

### 4.5.2 FBOD performance against various graph numbers $T$

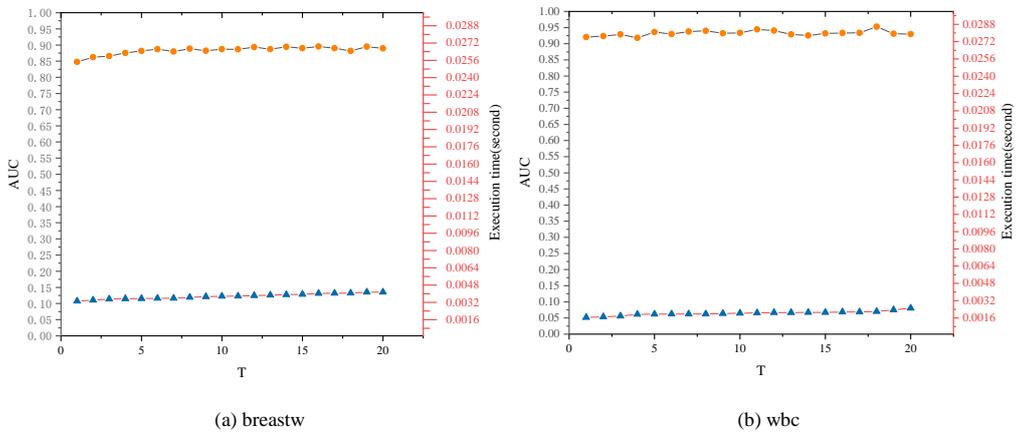

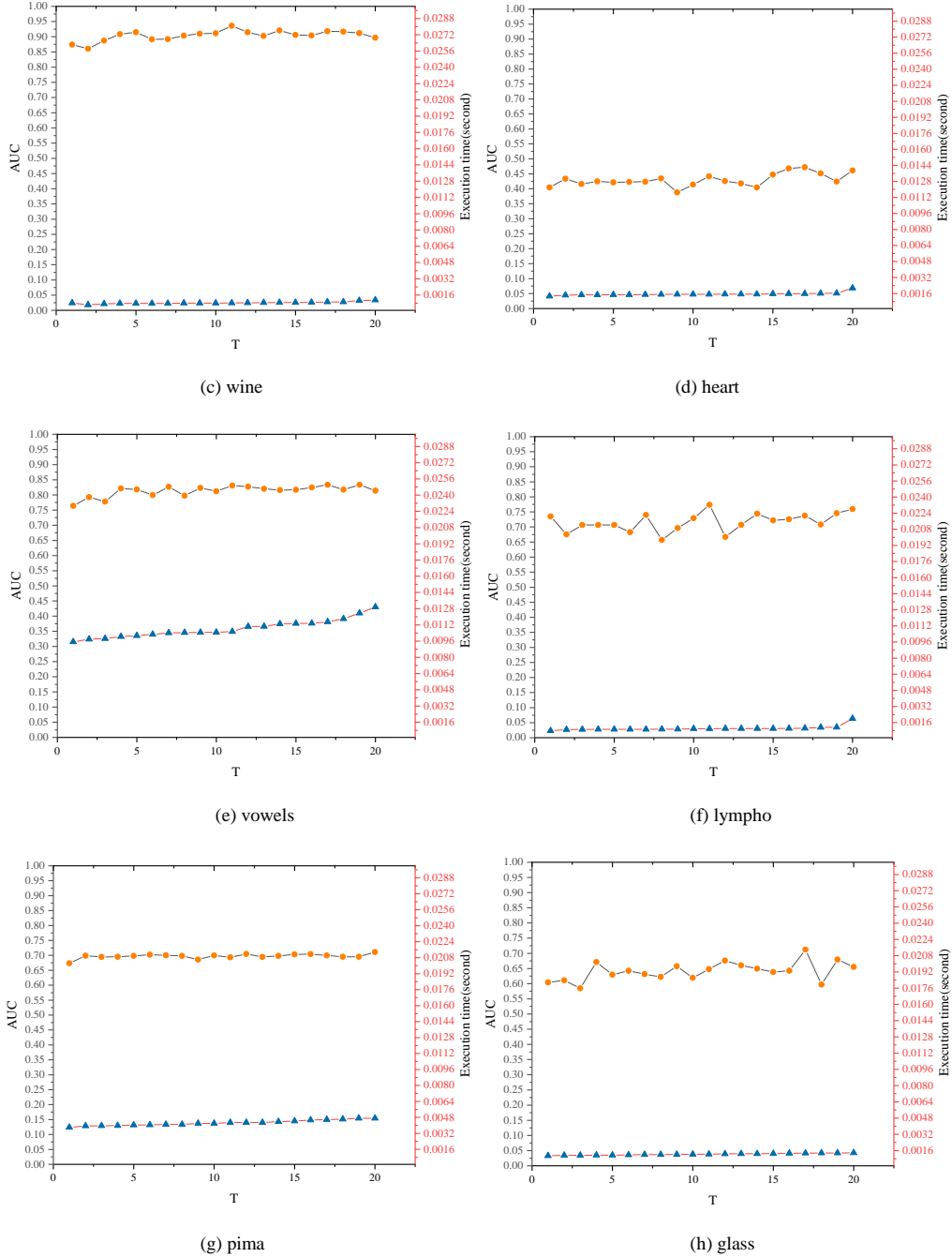

(c) wine  (d) heart  (e) vowels  (f) lympho  (g) pima  (h) glass

Fig. 11. FBOD AUC performance (y1-axis) and execution time (y2-axis) versus the different graph numbers (x-axis).

Parameter $T$ ranges from 1 to 20, and the value of $T$ increases by 1 each time ($k$=10). Fig 11 shows that when $T$ gradually increases, the AUCs of the four datasets of wine, heart, lympho, and glass are highly variable. The main reason for this phenomenon is that the distribution of normal objects in the dataset is too loose, which leads to a large variation in the selection of neighbors in each graph generation. Compared with the number of neighbors $k$, the parameter $T$ has less impact on the performance of the FBOD; its main role is to fine-tune the performance and improve the robustness of the algorithm.

## 5. Conclusion

In this paper, we proposed a fluctuation-based unsupervised outlier detection algorithm that changes the distribution of an original dataset by allowing objects to aggregate the feature values of their neighbors. Then, we define a new property of the object, fluctuation, in addition to distance, density and isolation. Finally, the fluctuation of the object is compared with its neighbors, and those objects with larger outlier factors are judged as outliers. FBOD is the first method that uses feature value propagation techniques and utilizes fluctuation for outlier detection. The results of experiments comparing FBOD with seven state-of-the-art algorithms on eight real-world tabular datasets show that FBOD achieves the best or next best AUC on six datasets. Meanwhile, FBOD achieves excellent detection results on video data. Most importantly, since the fluctuation-based algorithm does not need to calculate distance or density, the algorithm requires a very short execution time. It has high potential for real-world applications for outlier detection in large-scale data. Finally, we investigate the influence of hyperparameters in the FBOD on the detection effect in detail. In the future, we will attempt to take the FBOD algorithm by pretraining it to learn the fluctuation bounds of normal objects and then introduce it into the online application of data streams.

## Acknowledgments

This research was supported by the National Natural Science Foundation of China under grants 61862060, 61462079, 61562086, and 61562078.